\documentclass{llncs}
\usepackage{makeidx}  % allows for indexgeneration
\usepackage{graphics}
\usepackage{graphicx}
\usepackage{siunitx} 
\usepackage{amsmath}
\usepackage{amssymb}
\usepackage[caption=false]{subfig}

\newcommand{\figref}[1]{Fig.~\ref{#1}}
\newcommand{\secref}[1]{Sec.~\ref{#1}}

\newcommand{\tabref}[1]{Table~\ref{#1}}

\begin{document}
\frontmatter          % for the preliminaries
\pagestyle{headings} % switches on printing of running heads
\addtocmark{Accurate Pouring with an Autonomous
  Robot} % additional mark in the TOC
\title{Accurate Pouring with an Autonomous Robot Using an RGB-D
  Camera}
\titlerunning{Pouring Accuracy of an Autonomous Pouring System}  % abbreviated title (for running head)
%\author{Chau Do\inst{1} \and Wolfram Burgard\inst{1}}
\author{Chau Do \and Wolfram Burgard}
\authorrunning{Chau Do and Wolfram Burgard} % abbreviated author list (for running head)
\tocauthor{Chau Do, Wolfram Burgard}
\institute{University of Freiburg, Autonomous Intelligent Systems, Georges-K\"ohler-Allee 80, \\79110 Freiburg, Germany,
\email{\{do,burgard\}@informatik.uni-freiburg.de}}
%\\ WWW home page:
%\texttt{http://users/\homedir iekeland/web/welcome.html}
%\and
%Universit\'{e} de Paris-Sud,
%Laboratoire d'Analyse Num\'{e}rique, B\^{a}timent 425,\\
%F-91405 Orsay Cedex, France}

\maketitle              % typeset the title of the contribution

\begin{abstract}
  Robotic assistants in a home environment are expected to perform
  various complex tasks for their users. One particularly challenging
  task is pouring drinks into cups, which for successful
  completion, requires the detection and tracking of the liquid level during a
  pour to determine when to stop.  In this paper, we present a novel
  approach to autonomous pouring that tracks the liquid level using an
  RGB-D camera and adapts the rate of pouring based on the liquid
  level feedback. We thoroughly evaluate our system on various types
  of liquids and under different conditions, conducting over 250 pours
  with a PR2 robot.  The results demonstrate that our approach is able
  to pour liquids to a target height with an accuracy of a few
  millimeters.  
\end{abstract}
\section{Introduction}
A capable and effective domestic service robot must be able to handle
everyday tasks involving liquids.  Some examples are filling a cup or
measuring out a certain amount of liquid for baking or cooking.  This
requires the ability to perceive the liquid level while pouring and
using this information to decide when to stop pouring. This is a
challenging task, considering the large selection of liquids available
and their varying characteristics.  In this paper, we consider the
problem of tracking a liquid and determining when to stop pouring
using depth data from a low-cost, widely available RGB-D camera.

\begin{figure}[t]
	\centering \subfloat[][PR2 with cups and bottles used in the experiments]{
		\includegraphics[height=5cm]{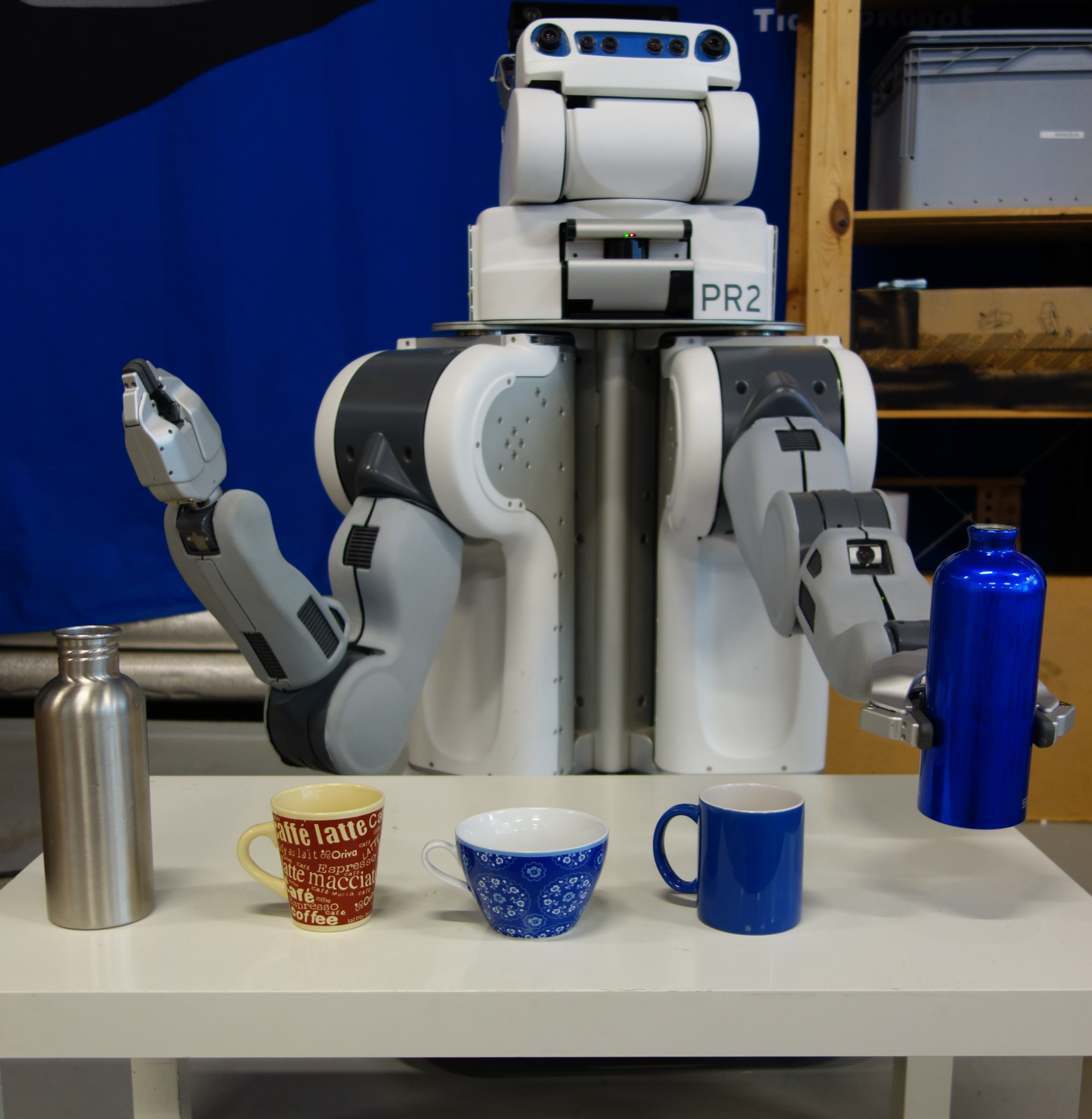}}
	\subfloat[][PR2 pouring water]{
		\includegraphics[height=5cm]{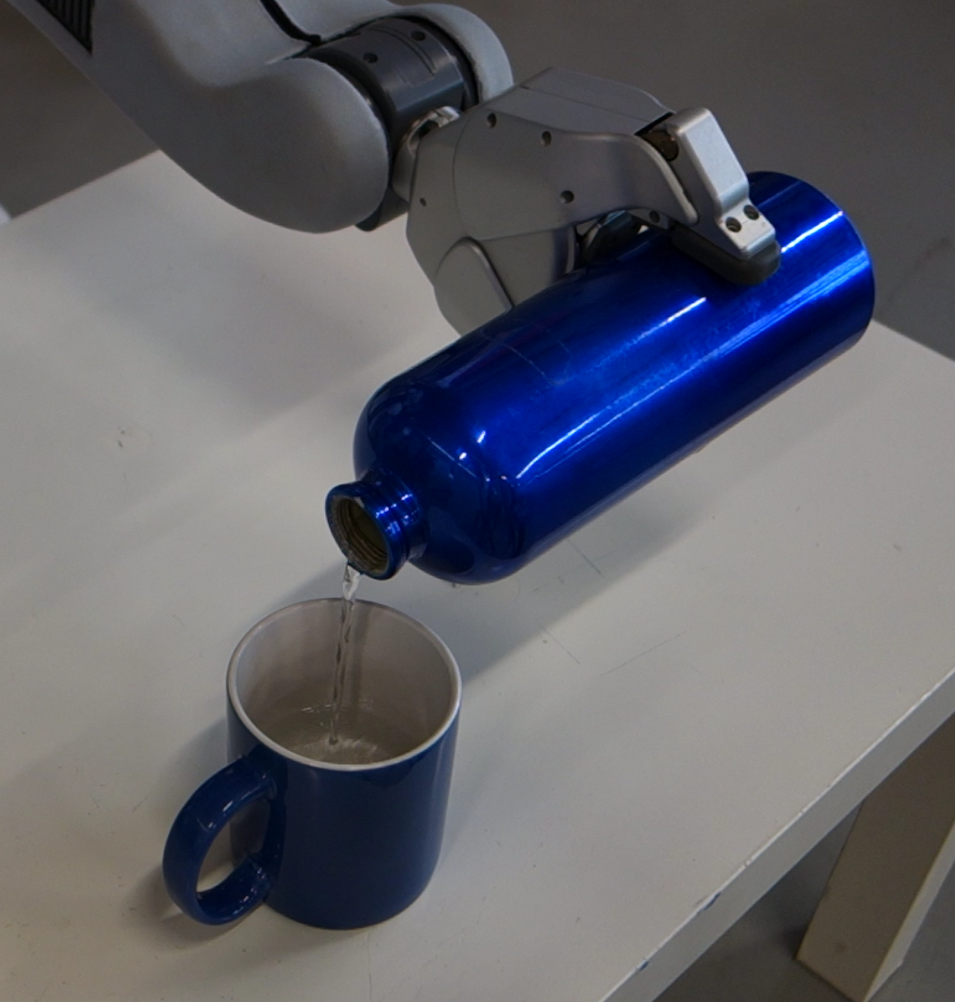}}
              \caption{PR2 robot with bottles and cups used in the
                experiments (a) and close-up of a water pouring trial
                (b).}\label{fig:robotic_setup}
\end{figure}

Visually, liquids change their appearance depending on the
environment, which makes it particularly challenging to detect the
fill level of a container using vision. In this paper we therefore
investigate the estimation of the fill level based on data provided by
a depth camera such as the ASUS Xtion Pro or Microsoft
Kinect. However, even for this modality the task at hand is
complicated by the fact that liquids have different appearances
depending on their transparency and index of refraction. For
transparent liquids such as water and olive oil, the infrared light is
\textit{refracted} at the liquid boundary, causing the resulting
liquid level to appear lower than it actually is. In the case of
water, depending on the view angle, a full cup appears one-third to
one-half full based on the depth data.  On the other hand, opaque
liquids such as milk and orange juice, reflect the infrared
light and the resulting depth measurement correctly represents the
real liquid height. The approach presented in this paper is designed
to deal with opaque and transparent liquids, by switching between the
detection approaches depending on the type of liquid. We utilize a
Kalman filter handling the uncertainty in the measurements and for
tracking the liquid heights during a pour. For controlling the pour,
we employ a variant of a PID controller that takes the perception
feedback into account. We demonstrate the effectiveness of our
approach through extensive experiments, which we conducted using a PR2
robot depicted in \figref{fig:robotic_setup}.

The two main contributions of this paper are a novel approach for
pouring liquids into a container up to a user-defined height and the
extensive analysis of the approach with respect to a large variety of
parameters of the overall problem.

\section{Related Work}
A substantial amount of research has been carried out on various aspects
of pouring. For example, Pan and Manocha~\cite{pan2016motion} and
Tamosiunaite et al.~\cite{tamosiunaite2011learning} focus on learning
a pouring trajectory, but do not consider the problem of pouring to a
specific height.  In the area of learning from humans, Langsfeld et
al.~\cite{langsfeld2014incorporating} and Rozo et
al.~\cite{rozo2013force} demonstrate how to pour a specific volume,
and learn the pouring motion parameters for that volume. However, in
their work no perception of the liquid is used.  Regarding the area
of liquid perception, Elbrechter et al.~\cite{elbrechter2015iros}
focus on the problem of detecting liquid viscosity.  Morris and
Kutulakos~\cite{morris2011pami} look at reconstructing a refractive
surface, but this requires a pattern placed underneath the liquid
surface. Mottaghi et al.~\cite{mottaghi2017see} use deep learning to
infer volume characteristics of containers.  Neither of these papers
consider the problem of tracking a liquid level and pouring to a
specific height. Yamaguchi et al.~\cite{yamaguchi2015pouring} focus on
the planning side of pouring.  They detect the fill level using a
plastic cup, modified such that the back half is colored and use color
segmentation. Our approach does not require modified containers.
Yamaguchi et al.~\cite{yamaguchi2016stereo} use a stereo camera and
apply optical flow detection to determine the liquid flow.  They do
not apply their approach to the problem of pouring to a specific
height. Furthermore, flow detection cannot be used to determine static
liquid already present in a container, which our approach is capable
of as shown by our experiments.  Yoshitaka et al.~\cite{hara2014iros}
use an RGB-D camera to detect the presence and height of liquid in a
cup.  For transparent liquids, they outlined a relationship between
the measured height from the depth data and the real liquid height.
They did not consider the problem of pouring liquids and only dealt
with the detection of static liquids.

In our previous work \cite{do2016iros}, we presented a probabilistic approach to liquid
height detection, which makes use of both RGB and depth data. 
It assumes no knowledge of the liquid type and only
considers a static liquid height (i.e., non-pouring) scenario.  This
previous approach uses multiple images and point clouds of the static
liquid, taken from different viewing angles. Capturing 
data from different viewing angles is required to disambiguate between liquid
types and accurately determine the height, since no prior knowledge of
the liquid is given.  In this paper, in contrast to our prior work, we
determine the height of the liquid during the pour and without
observing the cup from multiple viewpoints.

Schneck and Fox~\cite{schenck2016detection} demonstrate it is possible
to segment and track liquids using deep learning. However, this was
only shown with synthetic data.  In follow-up work, Schneck and
Fox~\cite{SchenckF16c} consider the problem of pouring to a specific
height. They train a deep network to classify image pixels as either
liquid or not liquid. For ground truth, they synchronized a thermal
camera with an RGB-D camera and recorded pouring data using water
heated to \SI{93}{\celsius}. Using this detection they then track the
liquid volume while pouring. For 30 pours, they reported a mean error
of \SI{38}{ml}. As we will demonstrate in our experimental
evaluations, our approach is able to achieve a substantially lower
mean error while only employing an RGB-D camera and not requiring
an expensive thermal camera as well as the heating of the liquid for
collecting the training data.

Compared to these previous approaches, the method presented in this
paper allows accurate pouring of liquids to given levels using a
low-cost RGB-D sensor.  It can be applied to different types of
liquids without any need for training and is usable in every
environment in which an RGB-D camera such as the Kinect or Xtion can
operate.

\section{The Autonomous Pouring System}

\begin{figure}
	\centering
	\includegraphics[width=1\linewidth]{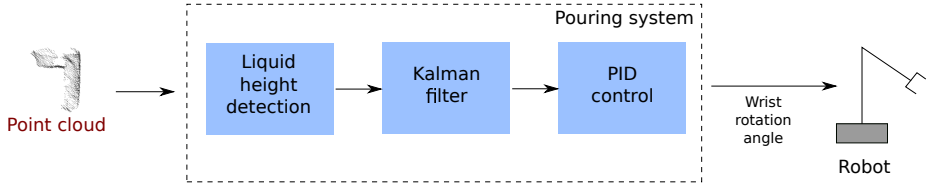}
	\caption{Overview of the approach: the sensor inputs are
          point clouds from an RGB-D camera. The output is a rotation
          angle for the wrist joint of the robot to pour the liquid.}
	\label{fig:system_overview}
\end{figure}

\figref{fig:system_overview} depicts our PR2 robot used for evaluating
our pouring approach. We control the pouring by rotating the angle of
the robot's wrist joint.  As input the system receives the point
clouds from the robot's RGB-D sensor. The cup should be positioned
such that the camera can look into it and can see part of the cup
bottom.

\subsection{Liquid Height Detection}

In our approach, the detection and tracking of the liquid level relies
only on the depth information from an RGB-D camera such as a Kinect or
an ASUS Xtion Pro. Throughout this paper, we assume that the robot
knows which liquid is being poured and which target height is
required. Typically, this information can easily be provided by the
user, for example, by issuing a command such as ``pour me a full cup
of water'' or following a recipe for baking and cooking.

There are two cases to consider with respect to the liquid type:
opaque and transparent liquids.  Opaque liquids, such as milk and
orange juice, are accurately represented in the point cloud and the
extracted height can be taken as the true liquid height.  Transparent
liquids such as water and olive oil, however, refract the light and
the point cloud height is incorrect. It is possible to estimate the
actual liquid height given a refracted depth measurement using a relationship 
based on the view angle and the index of refraction of the liquid (see Yoshitaka 
et al.~\cite{hara2014iros} and Do et al.~\cite{do2016iros} for a full derivation). 
This relationship for finding the liquid height $h$ is given by:
% To calculate the true liquid
%height $h$ for transparent liquids, our approach employs the following
%relationship:
\begin{equation}
\label{eq:noisefree}
h=\left(\frac{\sqrt{n_{l}^2 - 1 +\cos^2(\alpha)}}{\sqrt{n_{l}^2 - 1 +\cos^2(\alpha)}-\cos(\alpha)}\right)h_r.
\end{equation} 
Thus, we estimate the liquid height $h$ given the raw depth
measurement height $h_r$, index of refraction of the liquid $n_l$ and
the angle $\alpha$, where $h_r$ is the liquid height determined from
the point cloud and $\alpha$ is the incidence angle of infrared light
from the camera projector with respect to the normal of the liquid
surface. The latter is also determined from the point cloud.  The
index of refraction $n_l$ is determined from the liquid type, which is
provided by the user. It should be noted that Eq.~\eqref{eq:noisefree}
is an approximation and does not account for all physical effects.

To determine the liquid height from the depth data, we first need to
detect the cup.  We achieve this by finding the plane representing the
table and extracting all points above the table.  Then we use
RANSAC~\cite{fischler1981acm} to determine the cylinder model for the
cup.  In order to avoid detecting the cup rim as part of the liquid
height, we determine the diameter of the cup from the detected
cylinder model and only consider a reduced diameter section for the
liquid. In other words, we only search an area defined by the cup
height and the reduced diameter.  Finally we extract the points inside
this area and average over them to get the raw measured height.

\subsection{Kalman Filter}
\label{sec:kalman_filter}
To track the liquid height and filter the typically noisy depth
measurements from the RGB-D camera \cite{andersen2012kinect}, we use
the Kalman filter for tracking.  Thereby we assume that the liquid
height follows a constant velocity motion given by
\begin{equation}
x_k = Fx_{k-1} + w_k,
\end{equation}
where $w_k\thicksim\mathcal{N}(0,Q_k)$ is zero-mean Gaussian noise
with covariance $Q_k$, which represents the system noise and is given
by
\begin{equation}
Q_k = q
\begin{bmatrix}
\frac{1}{3}\Delta t^3 & \frac{1}{2}\Delta t^2 \\
\frac{1}{2}\Delta t^2 & \Delta t
\end{bmatrix}.
\end{equation}
Here, $x_k$ and $F$ are given by
\begin{align}
x_k  = \begin{bmatrix}
h_k \\
\dot{h}_k
\end{bmatrix} \mbox{and }
%x_k &= [h_k\ \dot{h}_k]^T, 
F &=  \begin{bmatrix}
1 & \Delta t \\
0 & 1
\end{bmatrix}.
\end{align}
The term $h_k$ refers to the liquid height and $\dot{h}_k$ refers to
the rate of change in liquid height at time $k$, while $\Delta t$ is the time interval.
Note that we chose not to model a system input, as this term and its
effect on the system depends on several factors such as the bottle
opening, amount of liquid in the bottle and the tilt angle of the
bottle.  Modeling this term from recorded data would be unique for
that situation only.  We determined the measurement noise covariance
matrix $R_k$ and $q$ from collected pouring data.

\subsection{Pouring Control}

To realize the pour, we need to determine the rotation angle of the
wrist joint. In our approach we employ a variant of the PID controller
that we augmented by additional control policies to ensure a smooth
pour. In the event that no liquid height is detected in the point
cloud, we slow down the pouring.  If no liquid height update is
received (i.e., the detection is too slow), we maintain the rotation
angle so as not to continue pouring blindly. As soon as we detect that
the required liquid height has been reached, we rotate the bottle back
to its initial position. 

\section{Experiments}
\label{sec:experiments}
\begin{figure}
	\centering
	\includegraphics[width=1\columnwidth]{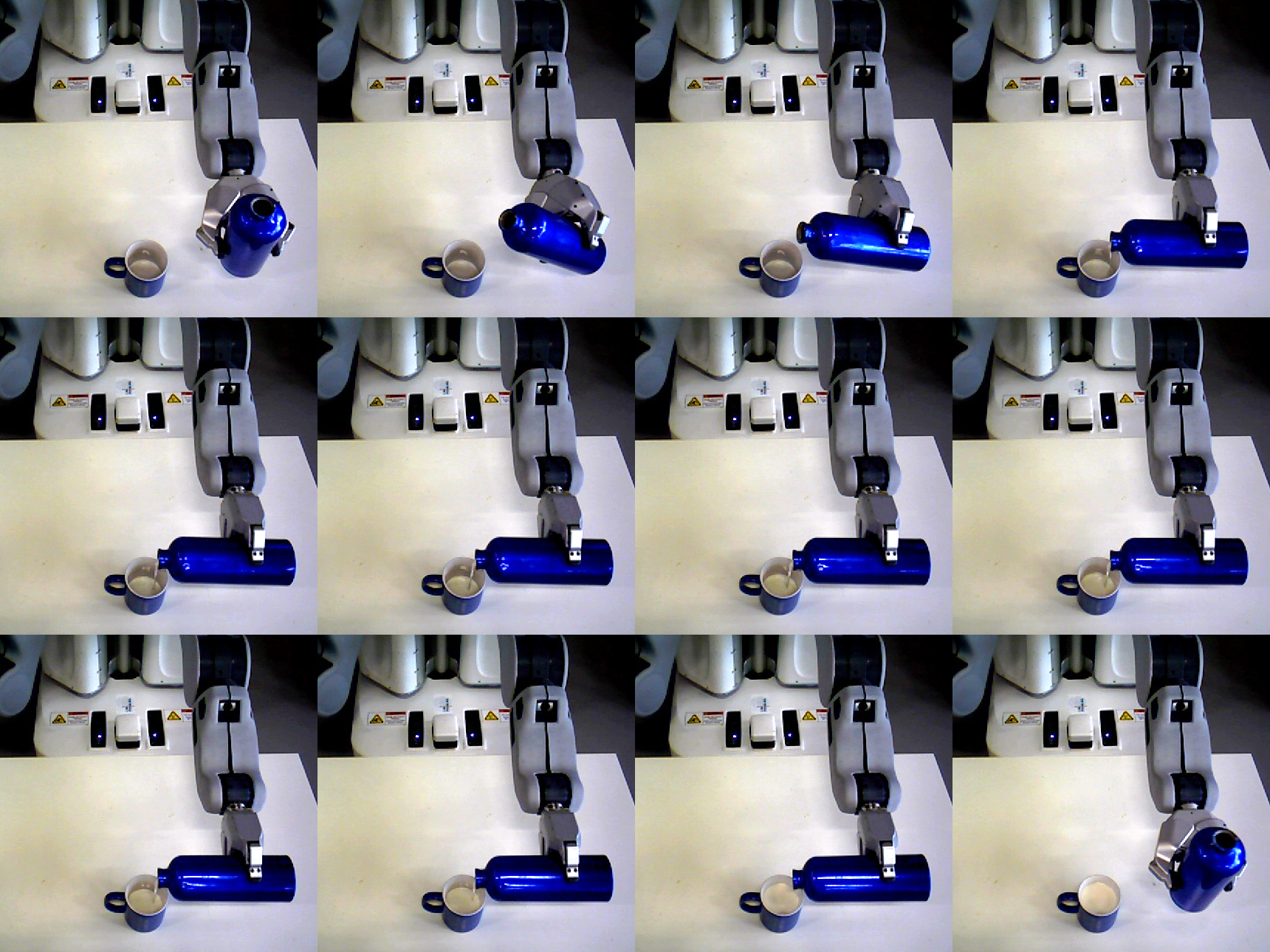}
	\caption{Pouring sequence for a milk pouring trail (best viewed enlarged).}
	\label{fig:pouring_sequence}
\end{figure}

We implemented our approach on a PR2 robot equipped with an ASUS Xtion
Pro camera and carried out a series of experiments to evaluate our
approach in real-world settings. The different cups and bottles
involved in the experiments can be seen in
\figref{fig:robotic_setup}. For the experiments, we positioned the cup
relative to the camera with its bottom center point at an approximate
horizontal distance of \SI{25}{cm} and an approximate vertical distance of
\SI{75}{cm}. The vertical angle of the camera was
chosen so that the cup bottom could be detected in the depth data. In
all experiments, we placed the bottle in the gripper of the PR2 and positioned it close to the cup.
Before the experiments, we fine-tuned the
parameters of our PID controller.  We found that proportional control
was sufficient for achieving accurate results. A pouring trial for milk
can be seen in \figref{fig:pouring_sequence}

In total, we performed six different types of experiments, which we
designed to analyze how changes in liquid type, initial liquid volume,
target height, bottles and cups affect the pouring accuracy. In all of
these experiments we used the same parameter values.  Overall, we
performed a total of 290 pours. To measure the ground truth data we
hand measured the liquid height with a ruler.

\subsection{Influence of Different Liquids}
\label{sec:diff_liquids}
To analyze the impact of the liquid type, we performed experiments
with water, carbonated water, olive oil, milk and orange juice.  The
first three are transparent liquids, while the latter two are opaque. 
Overall, they represent differences in properties such as 
viscosity, index of refraction and carbonation.

For each liquid we poured ten times. For each pour, the initial volume
in the bottle was randomly chosen from the values of [\SI{200}{ml},
\SI{250}{ml}, \SI{300}{ml}, \SI{350}{ml}, \SI{400}{ml}, \SI{450}{ml},
\SI{500}{ml}] and the target height in the cup was randomly chosen
from the range of [\SI{20}{mm}, \SI{30}{mm}, \SI{40}{mm}, \SI{50}{mm},
\SI{60}{mm}, \SI{70}{mm}].  We have the criteria that the initial
volume in the bottle should be at least \SI{100}{ml} more than the
target height, to make sure the robot does not pour all the liquid
out.  To focus only on the changes caused by the liquids, we used the
blue bottle and blue cup shown on the rightmost side of
\figref{fig:robotic_setup} for all the pours in this experiment.

\begin{figure}
	\centering
	\includegraphics[width=0.9\columnwidth]{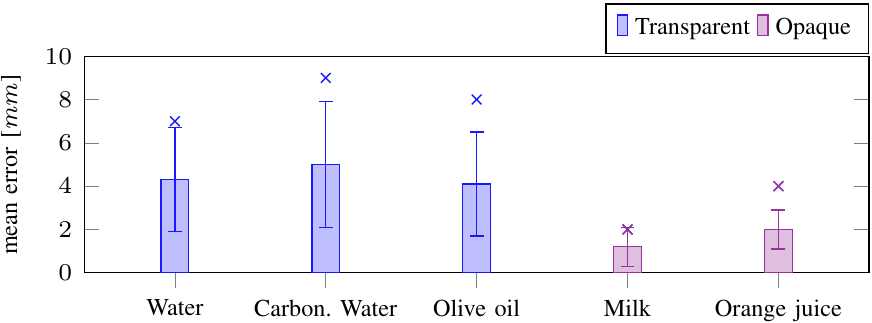}
	\caption{Mean error and standard deviation for ten pours
          of different liquids with varying initial volume and target
          height. The crosses mark the maximum error.}
	\label{fig:exp1_results}
\end{figure}

\figref{fig:exp1_results} shows the absolute mean error and standard
deviation for the ten pours of each liquid. The crosses show the
maximum error that occurred. In almost all cases the robot 
over-pours. We believe this is due to the fact that we send the \textit{STOP
  POUR} signal when we detect that the desired liquid height has been
reached. However, this is minimized by the PID controller, which reduces the 
amount poured as the liquid height approaches the target height. The opaque
liquid results show that this delay accounts for only around \SI{2}{mm} additional
liquid. Between the three transparent liquids there is not much variation and the same
can be said of the two opaque liquids.  

As can be seen, the transparent liquids have higher mean errors and
standard deviations versus the opaque ones. To investigate this
further, we additionally poured \SI{60}{mm} of water and milk into the
same blue cup and recorded 200 depth measurements. The results can be
seen in \tabref{table:static}. In this table, ``raw'' refers to the
raw depth measurement taken from the point cloud.  In the case of
water, this is before being transformed by Eq.~\eqref{eq:noisefree}.  For
the raw measurements, the standard deviation is about the same for
both water and milk. The transformed values for water have a higher
standard deviation. Looking at Eq.~\eqref{eq:noisefree}, it is clear that
any noise in the raw measurement $h_r$ is magnified by the term before
it. For the recorded data, the term before $h_r$ is around \SI{4.03}.
A second observation, is that the transformed value for water
underestimates the liquid height.  We note that Eq.~\eqref{eq:noisefree}
is only an approximation and does not factor in all physical
effects. Accordingly, we can expect a higher mean error and standard
deviation for the transparent liquids versus the opaque ones. 

\begin{table}[ht] \centering
	\caption{Static Error - Target Height \SI{60}{mm}} 
	\setlength{\tabcolsep}{5pt}
	\begin{tabular}{|c|c|}
		\hline
		%		\rule{0pt}{3ex}Target Height \SI{60}{mm} \\
		%		\hline
		\rule{0pt}{3ex} Measurement& $\mu\pm\sigma$ [mm]  \\
		\hline
		\rule{0pt}{3ex}Water (raw) &$18.64\pm 0.18$\\
		\rule{0pt}{3ex}Water &$58.77\pm 0.56$\\
		\rule{0pt}{3ex}Milk (raw) &$60.37\pm 0.12$\\
		\hline
	\end{tabular}
	\label{table:static}
\end{table}

\subsection{Influence of Varying Initial Volume}
In this experiment, we investigate how robust our system is to changes
in initial volumes in the bottle. We kept the target height
fixed at \SI{40}{mm} and varied the initial volume by changing it to one of
[\SI{350}{ml}, \SI{400}{ml}, \SI{450}{ml}, \SI{500}{ml}].  In
other words, for each volume we poured ten times to a height of
\SI{40}{mm}.  As before, we only used the blue bottle and blue cup.
The results for transparent liquids depicted in
\figref{fig:exp1_results} are very similar and the same can be said of
the opaque liquids. Hence for this experiment and the following ones
we only used water and milk.

\begin{figure}
	\centering
	\includegraphics[width=0.65\columnwidth]{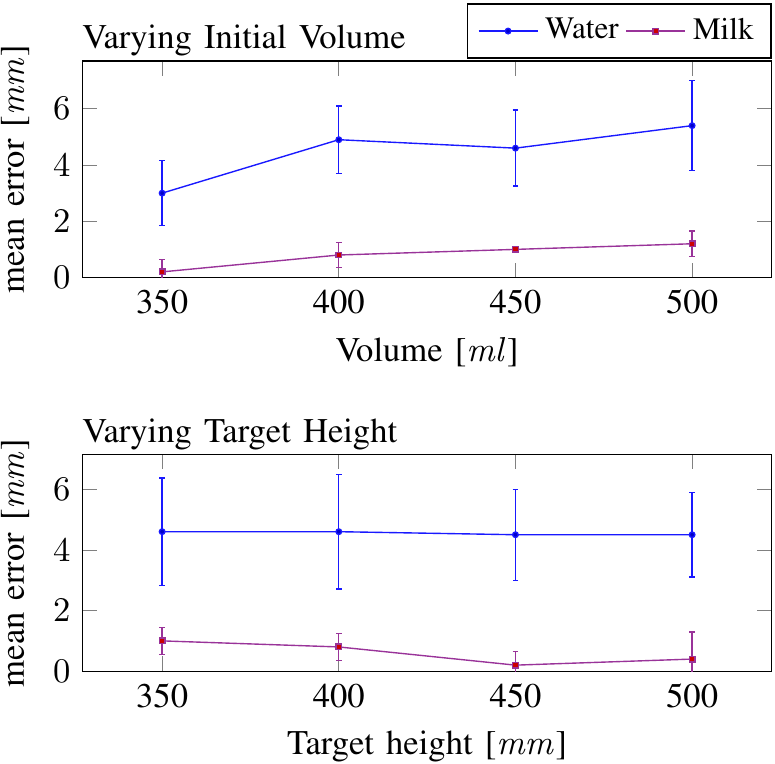}
	\caption{Mean error and standard deviation over ten pours for
          varying initial liquid volumes in the bottle and constant
          target height (top). Mean error and standard deviation over ten pours
	      for varying target height with constant initial liquid volumes (bottom).}
	\label{fig:exp2_exp3_results}
\end{figure}

The top figure in \figref{fig:exp2_exp3_results} shows the results of this experiment for the
pours.  There is a slight increase in mean error as the volume
increases. For water, the difference in mean error between
\SI{500}{ml} and \SI{350}{ml} is \SI{2.4}{mm} and for milk, it is
\SI{1}{mm}, which are not very large differences in view of the
overall results.

\subsection{Influence of Varying Target Heights}
Here we look at the influence of the target height. We kept the initial 
volume constant at \SI{400}{ml} and varied the 
target height to one of [\SI{30}{mm}, \SI{40}{mm}, \SI{50}{mm}, \SI{60}{mm}].
For each height, we poured ten times for both water and milk. The blue bottle
and blue cup were used each time for consistency. 

The results can be seen in the bottom figure of \figref{fig:exp2_exp3_results}. The mean errors and 
standard deviations remain fairly consistent for both liquids. So it does not appear
as if different target heights have a large influence on the pouring error.

%\begin{figure}
%	\centering
%	% \includegraphics[width=1\columnwidth]{pictures/setup1_cropped2.JPG}
%	\includegraphics[width=0.8\columnwidth]{figures/latex/exp3.pdf}
%	\caption{Mean error and standard deviation over ten pours for
%          different target heights with fixed liquid volume in the
%          bottle. }
%	\label{fig:exp3_results}
%\end{figure}

\subsection{Influence of Bottle Opening}
In this experiment, we investigate the influence of the bottle opening.
The silver bottle (referred to as the wide opening bottle) pictured on
the left in \figref{fig:robotic_setup} has an opening of \SI{4.5}{cm}
while the blue bottle (referred to as the small opening bottle),
pictured on the right, has an opening of \SI{2.5}{cm}. We conducted ten
pours each for water and milk, using the wide opening bottle. The
initial volumes and target heights were varied in the same manner as
in \secref{sec:diff_liquids} for each pour.

\begin{figure}  
	\centering
	\includegraphics[width=0.8\columnwidth]{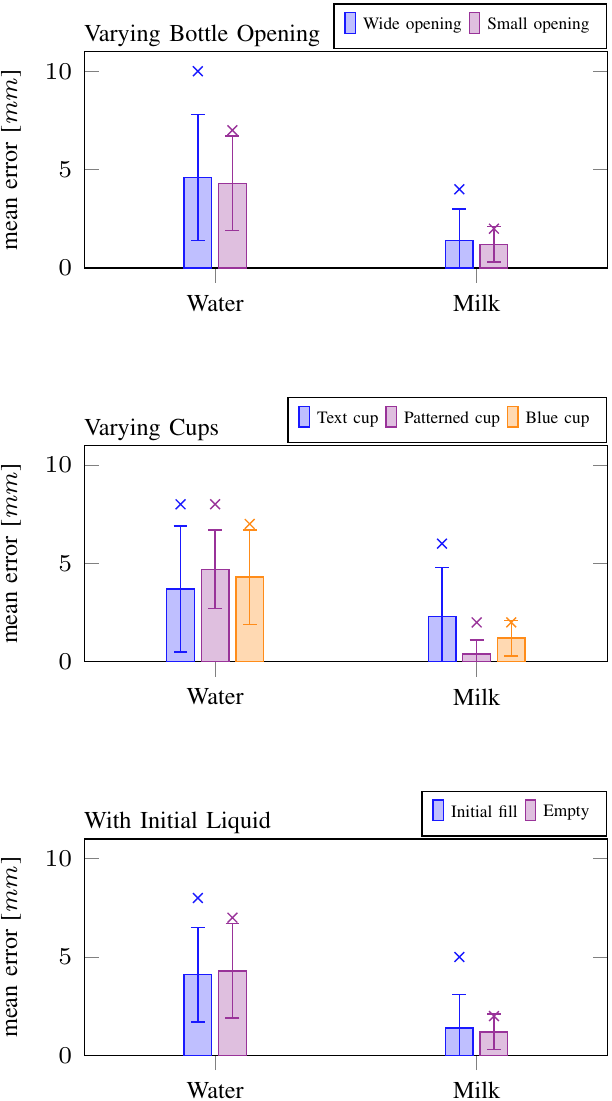}
	\caption{Mean error and standard deviation over ten pouring
          experiments with a wide and a small opening bottle (top). 
          Mean error and standard deviation over ten pours for
          each of the three different cups (middle). Mean error and standard deviation over ten pours for
          each liquid for pouring into a partly pre-filled cup (bottom). 
          In all cases, the crosses represent the worst results over all ten pours.}
	\label{fig:exp4_exp_5_exp7_results}
\end{figure}

The results can be seen in the top figure of \figref{fig:exp4_exp_5_exp7_results}. We include the
results for the small opening bottle from \secref{sec:diff_liquids}
for reference.  A small increase in mean error (\SI{0.3}{mm} for water
and \SI{0.2}{mm} for milk) for the wide opening bottle can be
seen. This can be expected since a wider opening makes it more
difficult to control the flow of the fluid as it leaves the bottle.
But overall, this experiment demonstrates that our system is able to
deal with different openings.

\subsection{Influence of Different Cups}
So far, each experiment was conducted with the blue cup, depicted on
the right side in \figref{fig:robotic_setup}. In this section we
investigate the influence of different cups, in particular whether the
shape and cup bottom diameter affect the system. We refer to the leftmost
cup in \figref{fig:robotic_setup} as the \textit{text cup}, the middle
cup as the \textit{patterned cup} and the rightmost cup as the
\textit{blue cup}.
For both the text cup and the patterned cup, we performed ten pours
each, using the blue bottle for water and milk. As in
\secref{sec:diff_liquids}, we varied the initial volume of liquid in
the bottle and the target height. We used the same approach for
detecting the cup in each case.

The results can be seen in the middle figure of \figref{fig:exp4_exp_5_exp7_results}. For reference we
include the results for the blue cup from
\secref{sec:diff_liquids}. In general, the differences in mean error
are minor between cups and there is no clear trend.  The cup with the
smallest bottom is the text cup, having a diameter of around
\SI{5}{cm}, compared to \SI{6}{cm} for the patterned cup and
\SI{7.5}{cm} for the blue cup. In the case of water, where it is the
refracted cup bottom that is being measured, one would expect a
smaller cup bottom would be more problematic, but this does not appear
to be the case. Furthermore, it was unclear beforehand whether the shape of the cups would affect Eq.~\eqref{eq:noisefree}
(due to reflections off the cup sides). But this does not appear to be an issue.

%\begin{figure} 
%	\centering
%	% \includegraphics[width=1\columnwidth]{pictures/setup1_cropped2.JPG}
%	\includegraphics[width=0.8\columnwidth]{figures/latex/exp5.pdf}
%	\caption{Mean error and standard deviation over ten pours for
%          each of the three different cups. The crosses represent the
%          worst results within the ten pours.}
%	\label{fig:exp5_results}
%\end{figure}

\subsection{Influence of Initial Liquid in Cup}
In this experiment, we look at how the system can deal with initial
liquid levels in the cup. For both water and milk, we performed ten
pours each.  In each pour, we chose an initial amount of liquid in the
cup ranging between \SI{10}-\SI{30}{mm}. We varied the initial volume
in the bottle and ensured the target height was at least \SI{10}{mm}
more than the initial liquid amount. Once again, we used the blue
bottle and blue cup.

The bottom figure in \figref{fig:exp4_exp_5_exp7_results} shows the mean errors, standard deviations
and maximum pour error.  We also include the case of starting with an
empty cup (see \secref{sec:diff_liquids} for reference).  The results
show only minor differences between pouring into an empty cup and
pouring into one with an initial amount of liquid. Accordingly our
system is able to robustly deal with situations in which the cup is
partly pre-filled.

\subsection{Comparison of Pouring Accuracy}
As noted before, Schneck and Fox~\cite{SchenckF16c} report a mean error of \SI{38}{ml} for water over 30 pours, using a system
that also controls the rotation angle of the wrist joint gripping the bottle. For each of our cups, we
take the worst 10-pour trial, which all occurred while pouring water, and converted the height errors to
volume errors. This resulted in mean volume errors of \SI{23.9}{ml} for the blue cup, \SI{13.2}{ml}
for the text cup and \SI{30.5}{ml} for the patterned cup. The patterned cup is also the widest cup of the three, meaning
errors in height result in larger volume errors.
Overall, our system was able to achieve better accuracy in pouring, while under more extreme testing conditions.

\section{Conclusion}
In this paper, we presented a novel approach for liquid pouring that
uses data received from an RGB-D camera and allows for pouring to a
specific, user-defined height. Our approach tracks the liquid height
during the pour which allows it to accurately stop as soon as the
desired height has been reached. We conducted extensive experiments
with our PR2 robot, which show that we are able to accurately pour
both a transparent and an opaque liquid under various conditions.  In
future work we will look at reducing the dependency on user input and
improving the handling of delays between
the perception and the controller so as to further improve the
accuracy.


%\clearpage
%\addtocmark[2]{Author Index} % additional numbered TOC entry
%\renewcommand{\indexname}{Author Index}
%\printindex
%\clearpage
%\addtocmark[2]{Subject Index} % additional numbered TOC entry
%\markboth{Subject Index}{Subject Index}
%\renewcommand{\indexname}{Subject Index}
%\input{subjidx.ind}
\end{document}